\documentclass{article}
\usepackage{graphicx} % Required for inserting images
\usepackage[utf8]{inputenc}
\usepackage[T1]{fontenc} %incluir em todos os papers em português
\usepackage[brazil]{babel}
\usepackage{hyperref}
\usepackage[top=2cm,bottom=2cm,left=2cm,right=2cm]{geometry}

\title{Experimentation, deployment and monitoring Machine Learning models: Approaches for applying MLOps}

\begin{document}

\maketitle

Diego Nogare \\
\textit{diego.nogare@gmail.com} \\
\textit{ORCID: 0000-0003-0796-9431} \\
{PPGEEC - Programa de Pós-Graduação em Engenharia Elétrica e Computação};
{ICTi - Instituto de Ciência e Tecnologia Itaú}
\newline
\newline
Ismar Frango Silveira \\
\textit{ismar.frango@gmail.com} \\
\textit{ORCID: 0000-0001-8029-072X} \\
{PPGEEC - Programa de Pós-Graduação em Engenharia Elétrica e Computação} \\
\newline

In recent years, Data Science has become increasingly relevant as a support tool for industry, significantly enhancing decision-making in a way never seen before. In this context, the MLOps discipline emerges as a solution to automate the life cycle of Machine Learning models, ranging from experimentation to monitoring in productive environments. Research results shows MLOps is a constantly evolving discipline, with challenges and solutions for integrating development and production environments, publishing models in production environments, and monitoring models throughout the end to end development lifecycle. This paper contributes to the understanding of MLOps techniques and their most diverse applications.

\section{Introduction}
In recent years, especially since 2010, Data Science has proven to be a fundamental discipline and a support tool for the industry to improve its decision-making supported by data. With the increased relevance of this area, the challenges of publishing the developed models into production to deliver the proposed value to end-users have become more prominent

To address these challenges, the MLOps discipline has proven to be a promising approach, enabling the automation and governance of the processes of experimenting, publishing and monitoring Machine Learning models. The creation of MLOps pipelines is one of the main strategies to ensure the effectiveness and efficiency of these processes.

This work is expected to contribute to the advancement of AI, promoting more efficient and transparent methodologies for end-to-end Machine Learning project development, looking for to answer the investigative question \textbf{"What are the main challenges faced by companies when publishing Machine Learning models into production, and how can the discipline of MLOps helps overcome them?"} and more specific questions like \textbf{"Why is it necessary to carry out continuous monitoring throughout the entire development lifecycle of machine learning models?"} and \textbf{"What are the essential steps to ensure an automated, efficient, and secure environment for publishing machine learning models?"}.

The remainder of the paper is organised as follow: in section 2 - MLOps pipeline, which explains the concepts and challenges of MLOps pipelines, in section 3 - Application and Case Study, applications and the benefits of implementing a solution with the stages of experimentation, publication and monitoring and three case studies from different fields of the industry that benefited from the implementation of MLOps are presented, and, in section 4 - Conclusion, the views of each of the three major areas explored are exposed.

\section{MLOps pipeline}
According to \cite{Subramanya2022} the key objectives of MLOps are to achieve faster development and deployment, mainly based on DevOps practices such as CI/CD, and to manage the ML lifecycle of models with high quality, reproducibility and end-to-end tracking. Additionally, MLOps enables a shorter code build-deploy cycle and aims to automate and monitor all ML steps. \cite{Markov2022} understands that the benefits of MLOps include automating the ML lifecycle, customizing models for product objectives, abstracting software infrastructure concerns, and increasing productivity. Research from \cite{Zhou2020} highlights the benefits of MLOps in uniting development and ML system operations, automating repetitive tasks like model training and code deployment, and enhancing model quality. This, in turn, reduces errors and downtime, enabling teams to detect and resolve problems quickly.
These points positively impact development teams by enabling them to work together more efficiently and collaboratively, which increases developer productivity by allowing data scientists and ML engineers to continuously monitor and adjust performance with increased traceability and transparency, so that teams can track changes to the model and understand how it is being used.

Following an attempt to answer the investigative questions, this paper presents details involving the steps for creating MLOps pipelines, and which elements are fundamental in strategies to develop data science projects, independent of the industry or field.

Considering that it is necessary to adopt clear and direct methodologies in the development stages of these pipelines, just as they are adopted in traditional software development, the stages of the model development life cycle will be discussed, which have been grouped into three key categories: Experimentation, Deployment and Model Monitoring.

\subsection{Experimentation}
The Experimentation stage of the Machine Learning model development life cycle involves technical aspects such as component development and testing, in addition to automated learning, which are widely researched in academic literature, as pointed by \cite{Ranawana2021, Gharibi2021, Testi2022}.

However, elements involving human factors such as collaboration in writing code, managing data science and software engineering work, dealing with deadlines and priorities, documenting interfaces and responsibilities, as well as planning, operating, and evolving the system are not so explored in publications, as highlighted by \cite{Fischer2020}.

Therefore, it is important that artificial intelligence projects are planned and integrated with other systems in a clear and direct way, following established standards such as CRISP-DM (Cross-industry standard process for data mining) or TDSP (Team data science process), such as pointed out by \cite{Niranjan2022, Prasad2022, Haakman2021}. This way, it is possible to guarantee the efficiency and quality of the development of Machine Learning models.

Iterative steps are part of the Machine Learning model lifecycle experimentation process and are executed repeatedly until a model achieves the expected result. These tasks are crucial to ensuring the efficiency and quality of model development, focusing on data preparation, model development and training, and validating its performance, as cited by \cite{Gharibi2021, Prasad2022, Rauschmayr2022, Isenko2022, Carqueja2022, Heuvel2020}.

Data preparation involves collecting, cleaning, processing, and transforming data, as well as selecting relevant features, and splitting data into training, validation, and test datasets. Model development and training involves choosing the architecture, defining hyperparameters, and applying supervised, unsupervised learning or reinforcement learning techniques. Model performance validation involves evaluating performance metrics, such as accuracy, precision, recall and F1-score, among others, as well as optimizing hyperparameters to improve model performance. Loop execution of these iterative steps is essential to ensure that the model matches business expectations and achieves good performance.

\subsection{Deployment}
Although publishing Machine Learning models is an important and challenging topic, it is not a subject frequently addressed in academic literature, as pointed out by \cite{Testi2022}.

However, the literature highlights the importance of ensuring that the model is implemented in a production environment in a safe and efficient way, following already established standards, as pointed out by \cite{Niranjan2022, Prasad2022}. Continuous monitoring after deployment must also be ensured in order to detect possible problems and ensure that the model is working correctly, as highlighted by \cite{Kannout2022}. The publish step can contain container creation and model serialization, as mentioned by \cite{Gharibi2021, Minon2022}.

There are several challenges to be addressed to publish a model efficiently and securely. Cientific literature highlights some of the main challenges, such as:

1. Variety of computing environments: One of the main challenges is the variety of computing environments in which the model can be deployed, such as an on-premises environment, in the cloud, in containers or at the edge, as mentioned by \cite{Minon2022}. Each environment may have its own limitations and requirements, which can make large-scale deployment of models difficult.

2. Programming languages and frameworks: Another challenge is the variety of programming languages and frameworks available to implement Machine Learning or Deep Learning models, as presented by \cite{Minon2022}. Each language and framework can have its own advantages and disadvantages, which can make it difficult to choose the best option for the model.

3. Portability: Portability is another important challenge, as the possibility of transporting the model publicated to other platforms may be limited, which affects the maintenance and portability of the model, as highlighted by \cite{Minon2022}.

4. Security: Security is a critical challenge as Machine Learning models can be vulnerable to malicious attacks such as data poisoning attacks or adversarial attacks, as pointed out by \cite{Gharibi2021}.

\subsection{Model Monitoring}
Continuous monitoring is a critical part of the MLOps pipeline lifecycle, as it helps identify and minimize impacts of data or concept drift, according to \cite{Rauschmayr2022, Kannout2022}.

ML models and data are constantly changing and this can happen unexpectedly, causing the model to become obsolete and respond with less accurate results, as \cite{Gharibi2021} explains. Continuous monitoring identify trends and patterns in data, which allows system adjustment to improve performance, as explained by \cite{Rauschmayr2022}.

Furthermore, continuous monitoring also involves quality assurance, management and maintenance of new models, as well as the specificity of security policies, protection and other non-functional requirements of the project, as highlighted by \cite{Carqueja2022}. Continuous monitoring also allows the identification of risks and maintenance of the model in production, aligned with the metrics established by the business, according to \cite{Kannout2022}. This is essential to ensure the efficiency and quality of the model in production, as well as to guarantee customer satisfaction.

The scientific papers highlights that what should be monitored depends on the specific context of the ML system in question and the relevant business metrics, as mentioned by \cite{Bourgais2022, Carqueja2022}. However, in general, the distribution of input and output data from the model must be monitored to detect data or concept drift, detection of missing values, outliers or extreme values and inconsistent data, without neglecting the identification of ethical or bias issues, as \cite{Matsui2022} highlights. Furthermore, it is important to monitor the model's performance in relation to performance evaluation metrics, such as accuracy, precision, recall, F1-score, AUC and ROC curve, among others, according to \cite{Carqueja2022}.

Another monitoring aspect that must be considered is the growing demand on issues related to model transparency seeking to offer better explainability, interpretability and accountability of AI, as mentioned by \cite{Heuvel2020, Tamburri2020}. Not to mention the importance of monitoring ethical and bias issues in Machine Learning models, as mentioned by \cite{Matsui2022}, because Machine Learning models can be vulnerable to discrimination and fairness issues that can negatively affect accuracy and effectiveness of the model, according to \cite{Bourgais2022}. This occurs, in large scale, because traditional models tend to operate as a black box, making it unlikely to have an effective understanding beyond control and evaluation of their operations, as well as in deep learning models that solve most problems by automatically learning from the input data, which also leads to difficulties in understanding and interpreting the models developed, according to \cite{Gharibi2021}.

\section{Application and Case Study}
According to \cite{Tchanjou2022} there is a lack of general knowledge and methodology that can facilitate and guide the adoption and migration of machine learning management tools by developers and data scientists, which impacts to increased overhead during configuration and maintenance of these environments and tools. \cite{Paleyes2022}, explains that MLOps can help address the challenges inherent in the process of improving the efficiency and effectiveness of deploying machine learning models. This could lead to greater adoption of ML in business processes and therefore greater potential benefits for the industry. According to \cite{Kannout2022}, developing and deploying MLOps pipelines helps ensure versatile model quality over an extended period of time, simultaneously driving accuracy and stability, reducing training time, and increasing model resilience. Furthermore, it mentions that machine learning pipelines are responsible for continuously monitoring and ensuring the quality of the models developed during their operation.

\subsection{Case Study}
Looking for of showing the MLOps applicability in different fields of the industry, three examples were selected from different fields that benefited from MLOps in their applications and reaped positive results.

\subsubsection{Financial Services}
Itaú Unibanco is the biggest financial bank in Latin America, currently has more than 65 million customers and has more than 70 PB of data. As explained by \cite{nogare2022}, the platform developed by Itaú helps simplify the management of Data Science models by enabling fast delivery, following good practices in software engineering, security and architecture of machine learning solutions integrated with the production process. Furthermore, the platform facilitates the observability of models and data so that retraining moments can be defined, as well as detecting deviations in statistical distributions, both of data and predictions, improving reliability for the business. An additional advantage is the greater transparency in the stages of publishing Data Science models in a production environment thanks to the governance required by these environments, such as change management and mandatory log storage. The proposed solution was created based on the main needs relating to an environment that supported the life cycle of Data Science models, including experimentation, deploy and monitoring, mapped together by the business and technology areas of Itaú Unibanco.

\subsubsection{Tech Solution Provider}
The EXPLAIN (EXPLanatory Interactive Artificial Intelligence for INdustry) project was led by \cite{FaubelSchmid2023} and built from interviews carried out with four project partner companies, each with experience related to the AI/ML sector. One to four employees from each company were interviewed, including data scientists/analysts, project managers, IT specialists, software engineers, and ML architects. A structured interview guide was developed and used to conduct a semi-structured interview, which covered all research questions. The information collected was analyzed and used to determine the extent to which MLOps is implemented by the project partner companies, what the MLOps architectures, tools and requirements are from the company perspective, and the requirements for developing a new MLOps software architecture. Interviews carried out with the project's partner companies showed that each of them uses MLOps differently, depending on the specific use case.

\subsubsection{Energy Consumption Prediction}
Research led by \cite{Fujii2023} presents a digital twin architecture for modeling energy consumption in homes. Through sensors implanted in devices, the solution collects device-level energy consumption data to recognize micro-moments and provide personalized, timed recommendations to users. The solution is more granular than the collaborative filtering-based approach, since each family cluster and its users are modeled in a specialized ontology. The solution is also robust to missing data and supports multiple time granularity from 1 to 1440 minutes (24 hours). Furthermore, data versioning is essential to ensure the reproducibility of experiments carried out in previous versions. MLOps was a fundamental part of the project, as it allowed the integration of the software development stages and operations of information technology systems, ensuring automation, monitoring and integration of tests and management of infrastructure as code, among other techniques, thus advancing towards the continuous delivery and deployment of the ML system.

\section{Conclusion}
Even though there is no concrete and objective definition of what MLOps is in the sense of defining it in a single sentence, its importance for the processes of publishing data science models is crystal clear, which can be observed from the research covered in this paper. It is possible to consider that the model development life cycle and all their stages are grouped into three key categories, namely, Experimentation, Deploy and Monitoring.

Experimentation is an iterative stage in the model life cycle, which involves evertything from data preparation, development and training of the model to validating its performance. This is a critical task in the model's life cycle, as it is in this phase that we look for to achieve the expected result and identify possible problems and challenges that will help answer the business problems.

Deploy Machine Learning models is a growing challenge in the area of Data Science and there is little academic literature on the subject. However, the stage of publishing Machine Learning models is involved with technical challenges, such as choosing the appropriate infrastructure to serve the model, configuring the production environment and integrating with other existing systems.

In terms of monitoring, it is important to monitor the models to avoid ethical and bias problems in Machine Learning projects, in order to minimize the risks of discrimination and fairness, which is mainly related to the transparency of the model, seeking to offer better explainability, interpretability and responsibility of the AI. Not to mention the operational monitoring that can monitor the detection of network failures, hardware failures and software errors, in addition to monitoring service interruptions, response latency and scalability of the solution, as well as attempted attacks that leave the model vulnerable.

Based on the steps grouped into these three key categories, it is expected that machine learning engineers can build environments that make it possible to take models from development to production in an automated, efficient and safe way, with continuous monitoring.

Details in each of the steps can be explored in future work in order to improve the processes.

\section{Acknowledgment}
We would like to express our deep gratitude to the Mackenzie Presbyterian Institute for the support provided during the development of this research. The financial support, infrastructure, and resources provided were fundamental to the success of this work.

We also want to extend our thanks to the Institute of Science and Technology Itaú (ICTi) for their continued encouragement and investment in Brazilian science. We firmly believe in the importance of their contribution to the advancement of knowledge and research in our country.

Any opinions, findings, and conclusions expressed in this manuscript are those of the authors and do not necessarily reflect the views, official policies or position of Itaú Unibanco.

\def\refname{References}
\bibliography{referencias}

\end{document}